\documentclass{article}
\usepackage{amsmath,graphicx}

\usepackage[preprint]{spconf}
\copyrightnotice{\copyright\ submitted to IEEE ICIP 2021.}

\title{Context-Aware Image Denoising with Auto-Threshold Canny Edge Detection to Suppress Adversarial Perturbation}

\name{Li-Yun Wang, Yeganeh Jalalpour, Wu-chi Feng}
\address{Department of Computer Science \\ 
         Portland State University, OR, USA \\
         email: \{liyuwang,yeganeh, wuchi\}@pdx.edu}

\usepackage{setspace}

\begin{document}
%
\maketitle

\begin{abstract}
This paper presents a novel context-aware image denoising algorithm that combines an adaptive image smoothing technique and color reduction techniques to remove perturbation from adversarial images. Adaptive image smoothing is achieved using auto-threshold canny edge detection to produce an accurate edge map used to produce a blurred image that preserves more edge features. The proposed algorithm then uses color reduction techniques to reconstruct the image using only a few representative colors. Through this technique, the algorithm can reduce the effects of adversarial perturbations on images. We also discuss experimental data on classification accuracy. Our results showed that the proposed approach reduces adversarial perturbation in adversarial attacks and increases the robustness of the deep convolutional neural network models.
\end{abstract}

\section{Introduction}
\label{sec:intro}

Researchers use deep convolutional neural networks (DCNNs) in different applications while dealing with data such as images, audios, and videos \cite{lecun2015deep}. There are various applications for processing and working with image data in the computer vision field that can be used in the real-world such as action recognition \cite{gochoo2017dcnn, shou2017cdc}, medical diagnosis \cite{xu2015empirical, araujo2017classification}, and self-driving vehicles \cite{bojarski2016end, rao2018deep}. 
However, adversarial attacks on DCNN models can negatively impact these applications' prediction results and make DCNN models less robust \cite{goodfellow14,moosavi16,Carlini17}. 
Image-related attacks are made by adding malicious and imperceptible perturbation to images and fool the DCNN models to classify images wrongly. For sensitive applications with the consequence of facing misclassifications, it would not be reliable to trust those models' output impacted by adversarial attacks.
Adversarial defense mechanisms allow DCNNs to counteract the effects of perturbation from adversarial attacks. Two primary approaches are preprocessing and adversarial training. Preprocessing techniques perform by preparing the data before passing it to the model, and these techniques are computationally effective \cite{ours,xu17,Dziugaite16}. The adversarial training techniques enhance the robustness of DCNNs by retraining DCNN models \cite{goodfellow2014explaining} or introducing new regularization terms \cite{cisse2017houdini,shaham2018understanding}. With these adversarial training techniques, DCNN models need many adversarial images as training images and a large amount of time to yield more robust models.

Similar to the methods proposed by Jalalpour {\it et al.} \cite{ours} and Feng {\it et al.} \cite{feng2020essential}, we leverage color reduction techniques to effectively remove perturbation noise. We propose using an adaptive Gaussian algorithm with auto-threshold canny edge detection to smooth out input images instead of using two thresholds to yield an edge map. This algorithm provides an accurate edge map to guide the AG algorithm using Gaussian kernels of various sizes. We then apply color reduction techniques to reduce the amount of color in the smoothed images. The final proposed algorithm can mitigate the adversarial effects and improve DCNN models' robustness in image classification tasks. This paper demonstrates the advantages of using intelligent parameter settings on the proposed algorithm. We empirically evaluated the three proposed algorithms, and the existing preprocessing approaches on an ImageNet-derived \cite{imagenet} dataset. Our experimental results indicate that the proposed algorithm considerably improves classification accuracy.

\section{Related Works}
\label{sec:rw}

\subsection{Adversarial Attack}


Some researchers use adversarial attacks to represent adversarial data can mislead DCNN models \cite{Carlini17, Taori18, Alzantot18}. 
Different attack approaches have been proposed to yield perturbed images with adversarial perturbation. Goodfellow {\it et al.} \cite{goodfellow14} proposed the fast gradient sign method (FGSM), which is efficient to create the perturbed images. FGSM maximizes the loss of classification problems for creating perturbation. Based on this idea, Kurakin {\it et al.} \cite{Kurakin16} proposed iterative FGSM (IFGSM), a refined version of the FGSM, to generate the adversarial images by iteratively applying FGSM. In each iteration, IFGSM introduces a small, fixed step size to modify the perturbation and accumulate it to create the consequent perturbated images. Moosavi {\it et al.} \cite{moosavi16} introduced an efficiently adversarial attack, DeepFool, by searching the nearest decision boundary and crossing it for perturbation generation. Carlini {\it et al.} also proposed an effective targeted and untargeted adversarial method, known as a C\&W attack \cite{Carlini17}. The authors modified the optimization problem by using various distance metrics to generate perturbed images.

\subsection{Adversarial Defense}
In this paper, we focus on image preprocessing techniques that strengthen the defense of DCNNs against adversarial perturbation. Researchers have shown that adversarial perturbation can be removed through preprocessing approaches. 
Xu {\it et al.} proposed using bit-depth reduction to reduce the number of bits in each color channel, thereby mitigating the effects of adversarial perturbation \cite{xu17}. They also used median and Gaussian smoothing to address the high-frequency variations typically found in adversarial images \cite{xu17}. Dziugaite {\it et al.} proposed the use of JPEG compression to counteract the effects of adversarial attacks \cite{Dziugaite16}. Jalalpour {\it et al.} proposed three color-reduction techniques to reduce the number of colors in adversarial images against adversarial attacks \cite{ours}, and Feng {\it et al.} proposed a context-aware image preprocessing approach that leverages a set of smoothing filters to mitigate the effects of adversarial perturbation \cite{feng2020essential}.

\section{The Proposed Algorithm}
\label{sec:approach}

We propose a context-aware image denoising algorithm with auto-threshold canny edge detection to remove the effect of perturbation in adversarial images. We begin by discussing our use of edge detection (Section \ref{edge}) to generate an edge map for later use. Section \ref{ag} describes applying the concept of adaptive Gaussian to our algorithm to produce blurred images with higher preservation of edge features. Finally, we describe our algorithm's use of color reduction (Section \ref{cr}) to reconstruct images using only a few representative colors.

\subsection{Edge Detection}\label{edge}

When creating an edge image for later use, we used an auto-threshold Canny edge detector \cite{lu2015cannylines} instead of a regular Canny edge detector \cite{canny1986computational} to strategically extract edge features without manually exploring the lower and upper thresholds in the Canny edge detection algorithm. Our justification is that the auto-threshold Canny edge detector sets the minimal meaningful and maximal meaningless gradient magnitude (edge intensity value) as the lower and upper thresholds, respectively. Therefore, the lower and upper thresholds can be automatically adjusted according to each image.

\begin{equation} \label{eq:nfa_equation}
\begin{gathered}
    NFA(S) = N_{p} \times P_{i}(u) \\
    N_{p} = \sum_{i=1}^{N_{h}} H(i) \times (H(i) - 1) \mathbin{/} 2
\end{gathered}
\end{equation}

The computation of the minimal meaningful and maximal meaningless gradient magnitude consists of three steps. First, the detector builds a gradient magnitude vector, $H(i), i = 1, ..., N_{h}$. Here, $i$ represents indices of elements in the vector. $N_{h}$ represents the maximum gradient magnitude. Each element represents a count of distinct gradient magnitude values, and these bins are used to compute a probability $P_{i}(u)$ with all the gradient magnitudes that are great than a constant (minimum gradient magnitude) $u$ for the $i$-th bin. The histogram and probabilities are used in the next step. Second, a Number of False Alarm (NFA) for an edge segment $S$ with a specific gradient magnitude is defined as equation \ref{eq:nfa_equation}. The NFA value represents the edge segment's expectation $S$ when an algorithm is run on a white-noise image. Finally, the detector applies the NFA value to define the minimal meaningful and maximal meaningless gradient magnitude. Because the NFA value is a probability, the detector applies a lambda value (vision-meaningful parameter) $\lambda_{v}$, which controls the lower bound of a noticeable gradient magnitude, to enhance the magnitude for visual effects.

\subsection{Adaptive Gaussian Algorithm}\label{ag}

Feng {\it et al.} \cite{feng2020essential} shows that the adaptive Gaussian smoothing approach effectively smooths out a considerable amount of perturbation and preserves much detail in new blurred images. Inspired by their work, we followed adaptive Gaussian's idea to smooth out the adversarial images in our algorithm. The algorithm we propose uses a parameter-free Canny edge detector to generate blurred edge maps, in contrast to the Sobel edge detection approach used by Feng {\it et al.} \cite{feng2020essential} or the regular Canny edge detector. Additionally, our proposed algorithm uses more than 3 Gaussian kernels with different kernel sizes instead of the fix 3 Gaussian kernels. Thus, our algorithm can handle various strength attacks. We used the parameter-free Canny edge detection because we only had one hyperparameter to adjust, allowing us to save time searching for the proper hyperparameter and generating a precise edge map. Our algorithm adds another hyperparameter, $\alpha$, to control the initial threshold value for the kernel size choice during the kernel selection process. This hyperparameter increases the flexibility of the kernel size choice. We select the $\alpha$ value from 1 to 6. We do not consider larger numbers because they generate too large an interval between two thresholds of determining the kernel sizes. Such an interval our proposed algorithm uses a same kernel to smooth out most regions. Our algorithm fails to take benefits of multiple kernels to generate the new blurred images with much better image quality.

\subsection{Color Reduction}\label{cr}

Jalalpour {\it et al.} \cite{ours} showed that color reduction techniques efficiently improves the robustness of the DCNN models. They reported that the color reduction techniques eliminate the least essential colors and reconstruct the image using only a few representative colors. 
Our proposed algorithm also uses the color reduction techniques to select {\it k} representative colors in an image. In particular, we adopted the idea of GK-means and fast GK-means proposed by Jalalpour {\it et al.} because these two approaches yield the overall best classification accuracy. We followed the same setting proposed by Jalalpour {\it et al.} for the {\it k} value. The algorithm we propose replaces fixed Gaussian smoothing with adaptive Gaussian smoothing with auto-threshold canny edge detection to preserve essential edge features and generate a more detailed image.

\vspace{-4mm}\section{Experiments}\label{sec:exp}

\subsection{Data set and Implementation Details}

To conduct our experimentation, we follow the same method of creating a validation set described by Jalalpour {\it et al.} \cite{ours} to match the DCNN models' inputs by selecting 2000 squarer images from the ImageNet Large Scale Visual Recognition Challenge (ILSVRC) validation set. We also followed the same procedure proposed by Jalalpour {\it et al.} \cite{ours} to generate adversarial images. For DCNN models, we considered Inception-v3 (IV3) and Resnet50 (R50) to perform image classification. We also adopted the same packages and tools mentioned by Jalalpour {\it et al.} \cite{ours} to implement existing image preprocessing techniques. We used OpenCV and NumPy to implement our proposed algorithm.

\subsection{Ablation Study}

\begin{figure}
    \centering
    \includegraphics[scale=0.4]{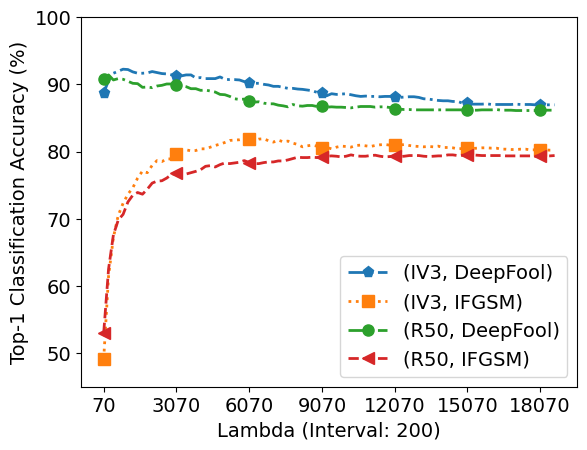}
    \caption{Top-1 classification accuracy versus $\lambda_{v}$ values. The first entry in the tuple represents the DCNN model. The second entry represents the adversarial attack. The Gaussian kernel sequence is 3, 5, 7, and 9.}
    \label{fig:lambda_val_figure}
\end{figure}

Figure \ref{fig:lambda_val_figure} shows the effect of different $\lambda_{v}$ values on top-1 classification accuracy for each adversarial attack and DCNN model. Within the small $\lambda_{v}$ values (e.g., 70–3070), the proposed adaptive Gaussian approach achieved high accuracy for DeepFool but not IFGSM. The difference in performance between adversarial attack methods was expected. For (IV3,DeepFool) and (R50, DeepFool), the top-1 classification accuracy decreased as $\lambda_{v}$ increased. For large $\lambda_{v}$ values (e.g., 12070-18070), our proposed approach does not have significant variation of top-1 classification accuracies. The algorithm only applies one of the Gaussian kernels to smooth out the entire perturbed image by referring to the edge map without soft edges. By doing this, the ultimate classification accuracies do not have significant changes, even if the proposed approach performs Gaussian smoothing depending on input images. In addition, the proposed algorithm achieved a higher top-1 accuracy against DeepFool attacks for small $\lambda_{v}$ values compared with IFGSM attacks. Lastly, the proposed adaptive Gaussian algorithm achieves high accuracy for large $\lambda_{v}$ values (e.g., 6070–9070). For DeepFool, we placed the point of highest accuracy at a $\lambda_{v}$ value of 670 because the top-1 classification accuracies for IV3 and R50 were 91.95\% and 90.85\%, respectively. For IFGSM, the optimal $\lambda_{v}$ value was 5870 because our proposed algorithm achieved 81.9\% and 78.65\% top-1 classification accuracy for IV3 and R50, respectively.

\begin{figure}
    \centering
    \includegraphics[scale=0.4]{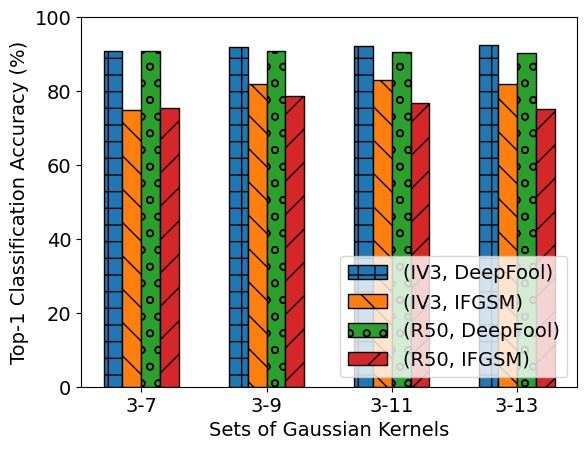}
    \caption{Top-1 classification accuracy versus sets of Gaussian kernels with various sizes. The first entry in the tuple represents a DCNN model. The second entry represents the adversarial attack. ``3-7" indicates kernel sizes 3, 5, and 7.}
    \label{fig:gaussian_size_figure}
\end{figure}

Figure \ref{fig:gaussian_size_figure} shows the classification accuracies for each of the four combinations of Gaussian kernels with varying kernel sizes. The sets of Gaussian kernels with distinct sizes (e.g., 3-9 and 3–13) have similar top-1 classification accuracy for DeepFool but not for IFGSM. In addition, for adversarial images generated by DeepFool using R50, the adaptive Gaussian algorithm that comprises auto-threshold canny edge detection with a large set of Gaussian kernels (e.g., 3–11 or 3–13) achieved a slightly lower top-1 classification accuracy compared with the small set of Gaussian kernels (e.g., 3–7 or 3-9). This result is attributable to DeepFool generating adversarial images efficiently. The adversarial perturbation generated by DeepFool is easy to be removed. The kernel with large kernel sizes smooths out some regions and discards essential features. Thus, the proposed algorithm with the large set of Gaussian kernels towards worse classification accuracy. For IFGSM, we choose 3–9 as the optimal combination of Gaussian kernels because this combination achieved the highest top-1 classification accuracy (e.g., the accuracies of the adversarial images generated by IV3 and R50 were 81.9\% and 78.65\%, respectively). For DeepFool, the algorithm using the 3–9 range achieved the highest top-1 accuracy. The classification accuracies for the adversarial images generated by IV3 and R50 were 91.95\% and 90.85\%, respectively.

\begin{figure}
    \centering
    \includegraphics[scale=0.4]{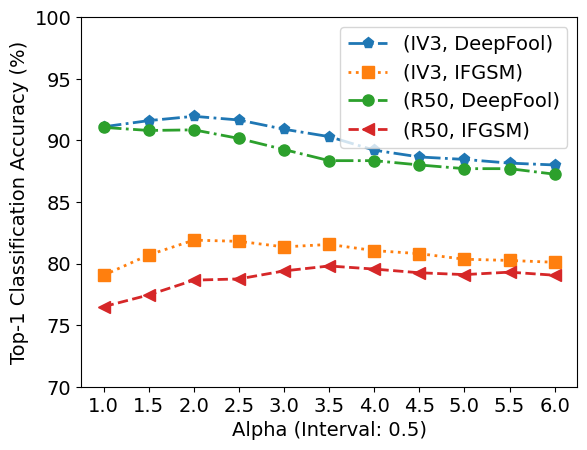}
    \caption{Top-1 classification accuracy versus alpha values. The first entry in the tuple represents the DCNN model. The second entry represents the adversarial attack. The Gaussian kernel sequence is 3, 5, 7, and 9. The $\lambda_{v}$ values are 670 and 5870 for DeepFool and IFGSM, respectively.}
    \label{fig:alpha_val_figure}
\end{figure}

Figure \ref{fig:alpha_val_figure} shows the top-1 classification accuracy for $\alpha$ values ranging from 1 to 6. An $\alpha$ value of 2 achieved the highest top-1 classification accuracy for DeepFool and IFGSM. This result is attributable to the fact that high and low $\alpha$ values produce corresponding initial thresholds in the linear function, generating inappropriate remaining thresholds. These values cause the proposed adaptive Gaussian algorithm to fail to select a proper Gaussian kernel through the linear function and achieve high classification accuracy. An $\alpha$ value of 2 achieved the highest top-1 classification accuracy for DeepFool and IFGSM. For DeepFool, the classification accuracies for the perturbed images generated by IV3 and R50 were 91.95\% and 90.85\%, respectively. For IFGSM, the classification accuracies of the perturbed images generated by IV3 and R50 were 81.90\% and 78.65\%, respectively.

\subsection{Results and Comparisons}

In this section, we present the effects of image preprocessing techniques on reducing adversarial images’ effects. Table 1 displays the empirical results. The approach we propose achieves the highest top-1 classification accuracy. The experimental results confirm our expectations that retaining subtle visual features and reducing the amount of color can mitigate perturbation effects in adversarial images. The adaptive Gaussian algorithm with auto-threshold canny edge detection achieved higher top-1 classification accuracy for the adversarial images generated by DeepFool using IV3 compared with existing image preprocessing techniques. Among our color reduction approaches, the proposed adaptive Gaussian method combined with the K-means method yields the lowest top-1 accuracy. This result is attributable to the fact that the adaptive Gaussian algorithm with auto-threshold canny edge detection combined with K-means leverages the larger Gaussian kernels to smooth out regions with a few edges. The blurred image inhibits the K-means’ selection of the most representative colors, and the image loses many essential visual details, resulting in low top-1 accuracy. For IV3, the proposed fast adaptive Gaussian method combined with the K-means method at 128 representative colors achieved 94.1\% and 88.45\% top-1 accuracy for DeepFool and IFGSM, respectively. This combination also achieved high top-1 accuracy for R50.

\begin{table}[h]
\centering
\scalebox{0.6}{
\begin{tabular}{c|c|c|c|c|}
\cline{2-5}
& \multicolumn{2}{c|}{\textbf{IV3}} & \multicolumn{2}{c|}{\textbf{R50}} \\ \cline{2-5}

& \multicolumn{1}{c|}{\textbf{DeepFool}} & \multicolumn{1}{c|}{\textbf{IFGSM}} & \multicolumn{1}{c|}{\textbf{DeepFool}} & \multicolumn{1}{c|}{\textbf{IFGSM}} \\ \cline{1-5}

\multicolumn{1}{|c|}{\begin{tabular}[c]{@{}c@{}} Perturbed image \end{tabular}} & 3.3\% & 0.85\% & 1.1\% & 0.15\% \\ \cline{1-5}
\multicolumn{1}{|c|}{\begin{tabular}[c]{@{}c@{}} Bit-depth reduction \\ (4 bits) \end{tabular}} & 73.3\% & 12.65\% & 77.85\% & 5.4\% \\ \cline{1-5}
\multicolumn{1}{|c|}{\begin{tabular}[c]{@{}c@{}} JPEG compression \\ (Level 75) \end{tabular}} & 88.45\% & 35.75\% & \textbf{91.35\%} & 38\% \\ \cline{1-5}
\multicolumn{1}{|c|}{\begin{tabular}[c]{@{}c@{}} Median smoothing \\ (Kernel size 3) \end{tabular}} & 79.2\% & 22.05\% & 80.5\% & 19.35\% \\ \cline{1-5}
\multicolumn{1}{|c|}{\begin{tabular}[c]{@{}c@{}} Gaussian smoothing \\ (Kernel size 5) \end{tabular}} & 88.3\% & 76.45\% & 88.7\% & 77.15\% \\ \cline{1-5}
\multicolumn{1}{|c|}{\begin{tabular}[c]{@{}c@{}} GK-means \\ (128 colors) \end{tabular}} & 91.5\% & 87\% & 85.2\% & 80.65\% \\ \cline{1-5}
\multicolumn{1}{|c|}{\begin{tabular}[c]{@{}c@{}} Fast GK-means \\ (128 colors) \end{tabular}} & 90.85\% & 86.95\% & 84.7\% & 80.1\% \\ \cline{1-5}
\hline\hline

\multicolumn{1}{|c|}{\begin{tabular}[c]{@{}c@{}}
Proposed adaptive Gaussian (ours)\\ ($\lambda_{v}$ 670, kernel size 3-9, $\alpha$ 2) \end{tabular}} & \textbf{91.95\%} & 70.5\% & 
\textbf{90.85\%} & 69.8\% \\ \cline{1-5}
\multicolumn{1}{|c|}{\begin{tabular}[c]{@{}c@{}} 
Proposed adaptive Gaussian (ours)\\ ($\lambda_{v}$ 5870, kernel size 3-9, $\alpha$ 2) \end{tabular}} & 90.4\% & \textbf{81.9\%} & 87.8\% & \textbf{78.65\%} \\ \cline{1-5}

\multicolumn{1}{|c|}{\begin{tabular}[c]{@{}c@{}}
Proposed adaptive Gaussian \\+ K-means (ours) \\ (128 colors, $\lambda_{v}$ 670) \end{tabular}} & 87.2\% & 80.1\% & 80.25\% & 67.8\% \\ \cline{1-5}
\multicolumn{1}{|c|}{\begin{tabular}[c]{@{}c@{}}
Proposed adaptive Gaussian \\+ K-means (ours) \\ (128 colors, $\lambda_{v}$ 5870) \end{tabular}} & 83.3\% & 78.8\% & 73.9\% & 65.6\% \\ \cline{1-5}
\multicolumn{1}{|c|}{\begin{tabular}[c]{@{}c@{}} Proposed fast adaptive Gaussian \\ + K-means (ours) \\ (128 colors, $\lambda_{v}$ 670) \end{tabular}} & \textbf{94.1\%} & \textbf{88.45\%} & 90.25\% & \textbf{81.9\%} \\ \cline{1-5}
\multicolumn{1}{|c|}{\begin{tabular}[c]{@{}c@{}} Proposed fast adaptive Gaussian \\ + K-means (ours)\\ (128 colors, $\lambda_{v}$ 5870) \end{tabular}} & 91.25\% & 87.45\% & 82.85\% & 78.7\% \\ \cline{1-5}
\end{tabular}}
\caption{Comparisons of different image preprocessing techniques for adversarial images generated by IV3 and R50. Entries refer to top-1 classification accuracy for defenses, and all results of the baseline techniques are the best for given appropriate hyperparameters.}
\label{top_1_accuracy_table}
\end{table}

\vspace{-4mm}
\section{Conclusion}
\label{sec:conclusion}
This paper proposes a new image preprocessing technique that effectively suppresses adversarial perturbation. We treated adversarial perturbation suppression as an image denoising problem, which we solved using context-aware image preprocessing with auto-threshold canny edge detection. Although our algorithm is simple, it achieved remarkable top-1 accuracy on the small ILSVRC data set we created and intelligently suppressed adversarial perturbation.

\bibliographystyle{IEEEbib}
\bibliography{icip2021}

\end{document}